%% file: ms.tex
\documentclass[10pt,twocolumn,letterpaper]{article}

\usepackage{iccv}
\usepackage{times}
\usepackage{epsfig}
\usepackage{graphicx}
\usepackage{amsmath}
\usepackage{amssymb}

\input{dg_usepackage}
\input{dg_def}

\def\redtext{\textcolor{red}}

\iccvfinalcopy 

\begin{document}

\title{Memorizing Normality to Detect Anomaly: Memory-augmented Deep Autoencoder for Unsupervised Anomaly Detection}

\author{Dong Gong{$^{1}$}, Lingqiao Liu{$^{1}$}, Vuong Le{$^{2}$}, Budhaditya Saha{$^{2}$},\\ Moussa Reda Mansour{$^{3}$}, Svetha Venkatesh{$^{2}$}, Anton van den Hengel{$^{1}$}
\\
$\!\!^{1}$The University of Adelaide, Australia~~ $^{2}$A2I2, Deakin University~~ $^{3}$University of Western Australia\\
{\tt\small \url{https://donggong1.github.io/anomdec-memae}}
}
 
\maketitle

\begin{abstract}
Deep autoencoder has been extensively used for anomaly detection. Training on the normal data, the autoencoder is expected to produce higher reconstruction error for the abnormal inputs than the normal ones, which is adopted as a criterion for identifying anomalies. However, this assumption does not always hold in practice. It has been observed that sometimes the autoencoder ``generalizes'' so well that it can also reconstruct anomalies well, leading to the miss detection of anomalies.  To mitigate this drawback for autoencoder based anomaly detector, we propose to augment the autoencoder with a memory module and develop an improved autoencoder called memory-augmented autoencoder, i.e. MemAE.  Given an input, MemAE firstly obtains the encoding from the encoder and then uses it as a query to retrieve the most relevant memory items for reconstruction. At the training stage, the memory contents are updated and are encouraged to represent the prototypical elements of the normal data. At the test stage, the learned memory will be fixed, and the reconstruction is obtained from a few selected memory records of the normal data. The reconstruction will thus tend to be close to a normal sample. Thus the reconstructed errors on anomalies will be strengthened for anomaly detection. MemAE is free of assumptions on the data type and thus general to be applied to different tasks. Experiments on various datasets prove the excellent generalization and high effectiveness of the proposed MemAE. 
\end{abstract}

\section{Introduction}
 Anomaly detection is an essential task with critical applications in various areas, such as video surveillance \cite{luo2017revisit}. 
The unsupervised anomaly detection \cite{zimek2012survey,zhai2016deep,zong2018deep,sabokrou2018adversarially,golan2018deep} is to learn a normal profile given only the normal data examples and then identify the samples not conforming to the normal profile as anomalies, which is challenging due to the lack of human supervision. 
Notably, the problem becomes even more difficult when the data points lay in a high-dimensional space (\ie videos), since modeling the high-dimensional data is notoriously challenging \cite{zimek2012survey}.

Deep autoencoder (AE) \cite{bengio2007greedy,kingma2013auto} is a powerful tool to model the high-dimensional data in the unsupervised setting. It consists of an encoder to obtain a compressed encoding from the input and a decoder that can reconstruct the data from the encoding. The encoding essentially acts as an information bottleneck which forces the network to extract the typical patterns of high-dimensional data. 
In the context of anomaly detection, the AE is usually trained by minimizing the reconstruction error on the normal data and then uses the reconstruction error as an indicator of anomalies. It is generally assumed \cite{zong2018deep,hasan2016learning,zhao2017spatio} that the reconstruction error will be lower for the normal input since they are close to the training data, while the reconstruction error becomes higher for the abnormal input.

\begin{figure}[!t]
\centering
\includegraphics[trim =0mm 0mm 0mm 0mm, clip, width=1\linewidth]{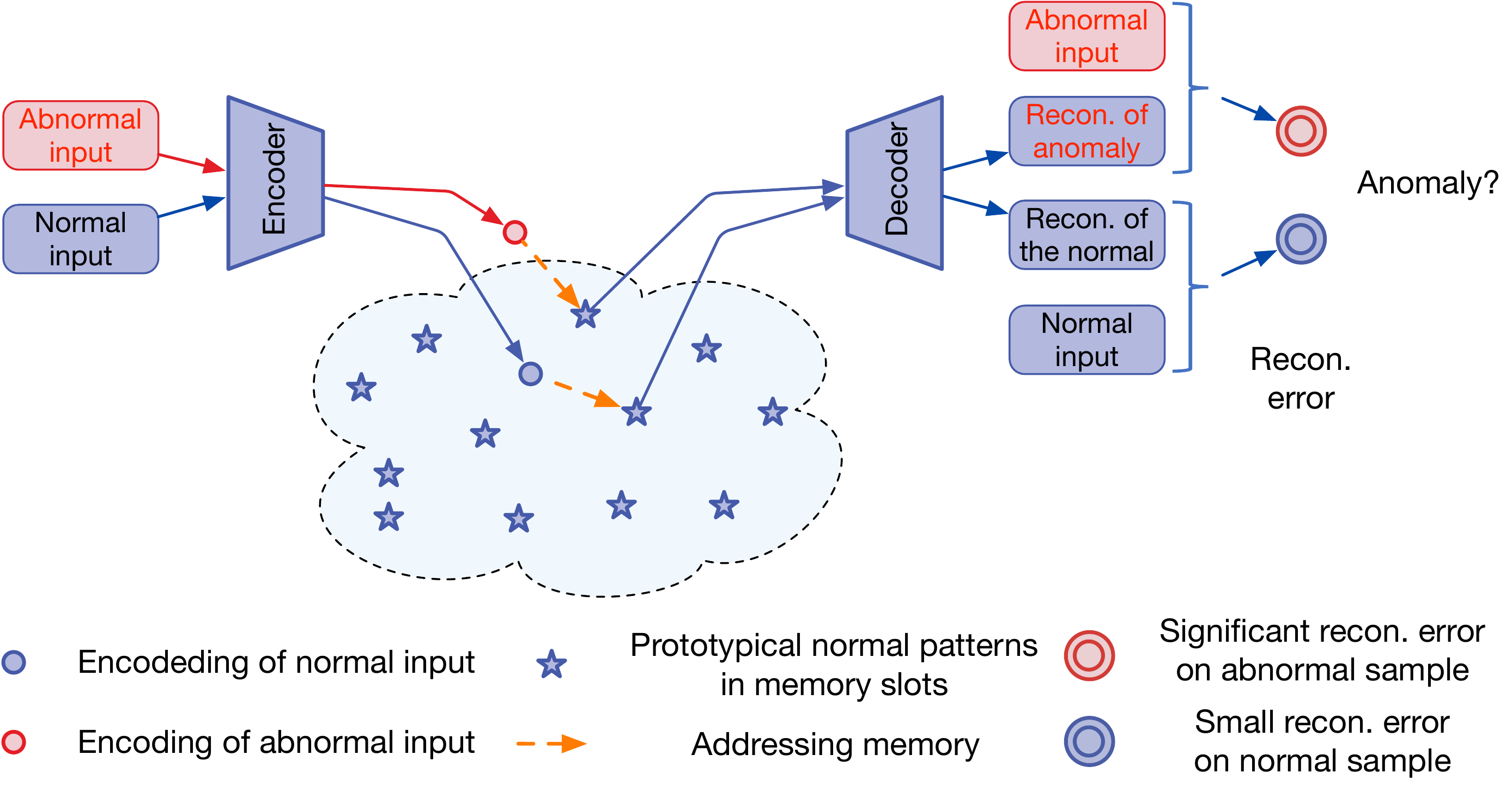}
\caption{Anomaly detection via the proposed MemAE. 
After training on the dataset only with normal samples, the memory in MemAE records the prototypical {normal patterns}. 
Given an \emph{abnormal} input, MemAE retrieves the most relevant \emph{normal} patterns in memory for reconstruction, resulting in an output significantly different to the abnormal input. To simplify the visualization, we assume only one memory item is addressed here. }
\label{fig:memae_diag}
\vspace{-0.4cm}
\end{figure}

However, this assumption may not always hold, and sometimes the AE can ``generalize'' so well that it can also reconstruct the abnormal inputs well. This observation has been made in the existing literature \cite[Figure 1]{zong2018deep} and also in this paper (See Figure \ref{fig:mnist_vis} and \ref{fig:recon_error_map}). 
The assumption that anomaly incurs higher reconstruction error might be somehow questionable since there are no training samples for anomalies and the reconstruction behavior for anomaly inputs should be unpredictable. 
If some anomalies share common compositional patterns (\eg local edges in images) with the normal training data or the decoder is ``too strong'' for decoding some abnormal encodings well, AE is very likely to reconstruct the anomalies well.

To mitigate the drawback of AEs, we propose to augment the deep autoencoder with a memory module and introduce a new model memory-augmented autoencoder, \ie MemAE. 
Given an input, MemAE does not directly feed its encoding into the decoder but uses it as a query to retrieve the most relevant items in the memory. 
Those items are then aggregated and delivered to the decoder. 
Specifically, the above process is realized by using attention based memory addressing. 
We further propose to use a differentiable hard shrinkage operator to induce sparsity of the memory addressing weights, which implicitly encourage the memory items to be close to the query in the feature space.

\par
In the training phase of MemAE, we update the memory content together with the encoder and decoder. Due to the sparse addressing strategy, the MemAE model is encouraged to optimally and efficient use the limited number of memory slots, making the memory to record the prototypical normal patterns in the normal training data to obtain low average reconstruction error (See Figure \ref{fig:vis_mem_sampling}). 
In the test phase, the learned memory content is fixed, and the reconstruction will be obtained by using a small number of the \emph{normal memory items}, which are selected as the neighborhoods of the encoding of the input. 
Because the reconstruction is obtained normal patterns in memory, it tends to be close to the normal data. Consequently, the reconstruction error tends to be highlighted if the input is not similar to normal data, that is, an anomaly. 
The schematic illustration is shown in Figure \ref{fig:memae_diag}. 
The proposed MemAE is free of the assumption on the types of data and thus can be generally applied to solve different tasks. We apply the proposed MemAE on various public anomaly detection datasets from different applications. Extensive experiments prove the excellent generalization and high effectiveness of MemAE.

\section{Related Work}

\noindent\textbf{Anomaly detection}
In unsupervised anomaly detection, only normal samples are available as training data \cite{chandola2009anomaly}. 
A natural choice for handling the problem is thus the one-class classification methods, such as one-class SVM \cite{chen2001one,scholkopf2001learning} and deep one-class networks \cite{ruff2018deep,chalapathy2018anomaly}, which seeks to learn a discriminative hyperplane surrounding the normal samples. 
Unsupervised clustering methods, such as the k-means method and Gaussian Mixture Models (GMM) \cite{zimek2012survey,xiong2011group}, have also been applied to build a detailed profile of the normal data for identifying the anomalies. 
These methods usually suffer from suboptimal performance when processing high-dimensional data.

\par
\emph{Reconstruction-based methods} are proposed relying on an assumption that the anomalies cannot be represented and reconstructed accurately by a model learned only on normal data \cite{zong2018deep}. 
Different techniques, such as PCA methods \cite{jolliffe2011principal,kim2009observe} and sparse representation \cite{lu2013abnormal,zhao2017spatio}, have been used to learn the representation of the normal patterns. Specifically, sparse representation methods \cite{lu2013abnormal,zhao2017spatio} jointly learn a dictionary and the sparse representation of the normal data for detecting anomalies. The restricted feature representations limit the performances. 
Some very recent works \cite{zhai2016deep,zhou2017anomaly,zong2018deep,chong2017abnormal} train deep autoencoders for anomaly detection. 
For example, 
structured energy based deep neural network \cite{zhai2016deep} is used to model the training samples. Zong \etal \cite{zong2018deep} proposed to jointly model the encoded features and the reconstruction error in a deep autoencoder. 
Although the reconstruction based methods have achieved fruitful results, their performances are restricted by the under-designed representation of the latent space.

\par
Due to the critical application scenario, a series of methods are specifically designed for \emph{video anomaly detection} \cite{liu2017future,zhao2011online,hasan2016learning,leyva2017lv}. 
Kim and Grauman \cite{kim2009observe} use a mixture of probabilistic PCA (MPPCA) to model optical flow features. Mahadevan \etal \cite{mahadevan2010anomaly} model the video via a mixture of dynamic textures (MDT). Lu \etal \cite{lu2013abnormal} proposed an efficient sparse coding-based method with multiple dictionaries. 
Zhao \etal \cite{zhao2011online} update the dictionary in an online manner. 
Deep learning based methods \cite{hasan2016learning,luo2017revisit,liu2017future,sabokrou2018adversarially} are proposed to use the information in both the spatial and temporal domain. 
Hasan \etal \cite{hasan2016learning} detect the anomalies according to the reconstruction error of a convolutional AE. 
Zhao \etal \cite{zhao2017spatio} proposed to use 3D convolution based reconstruction and prediction. 
Luo \etal \cite{luo2017revisit} iteratively update the sparse coefficients via a stacked RNN to detect anomalies in videos. 
Liu \etal \cite{liu2017future} train a frame prediction network by incorporating different techniques including gradient loss, optical flow, and adversarial training. However, these methods lack a reliable mechanism to encourage the model to induce large reconstruction error on the anomalies.

\noindent\textbf{Memory networks}
Memory-augmented networks have attracted increasing interest for solving different problems \cite{graves2014neural,WestonCB14memnet,santoro2016one}. 
Graves \etal \cite{graves2014neural} use external memory to extend the capability of neural networks, in which content-based attention is used for addressing the memory. 
 Considering that memory can record information stably, Santoro \etal \cite{santoro2016one} use a memory network to handle the one-shot learning problem. The external memory has also been used for multi-modal data generation \cite{kim2018memorization,li2016learning}, for circumventing the mode collapse issue and preserving detailed data structure.

\begin{figure*}[!t]
\centering
\includegraphics[trim =0mm 2mm 0mm 0mm, clip, width=0.8\linewidth]{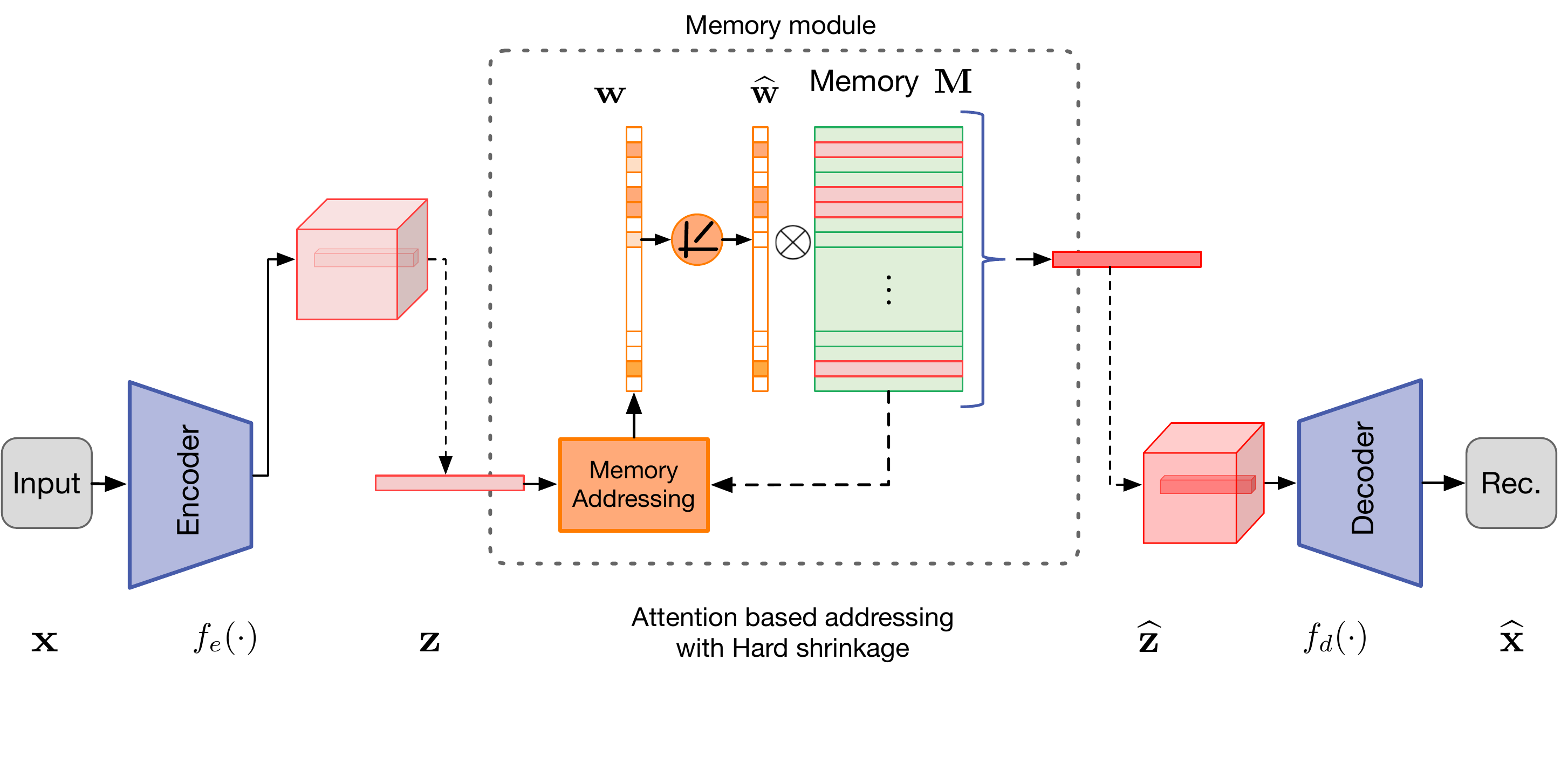}
\vspace{-0.6cm}
\caption{Diagram of the proposed MemAE. The memory addressing unit takes the encoding $\bz$ as query to obtain the soft addressing weights. The memory slots can be used to model the whole encoding or the features on one pixel of the encoding (as shown in the figure). Note that $\hatbw$ is normalized after the hard shrinkage operation. }
\label{fig:memae_stru}
\vspace{-0.2cm}
\end{figure*}

\section{Memory-augmented Autoencoder}

\subsection{Overview}

The proposed MemAE model consists of three major components - an encoder (for encoding input and generating query), a decoder (for reconstruction) and a memory module (with a memory and the associated memory addressing operator). As shown in Figure \ref{fig:memae_stru}, given an input, the encoder first obtains the encoding of the input.
By using the encoded representation as a query, the memory module retrieves the most relevant items in the memory via the attention-based addressing operator, which are then delivered to the decoder for reconstruction. 
During training, the encoder and decoder are optimized to minimize the reconstruction error. The memory contents are simultaneously updated to record the prototypical elements of the encoded normal data. 
Given a testing sample, the model performs reconstruction merely using a restricted number of the normal patterns recorded in the memory. 
As a result, the reconstruction tends to be close to the normal sample, resulting in small reconstruction errors for normal samples and large errors on anomalies, which will be used as a criterion to detect the anomalies.

\subsection{Encoder and Decoder}
The encoder is used to 
represent the input in an informative latent domain. The encoded representation performs as a query to retrieve the relevant items in the memory. In our model, the encoder can be seen as a \emph{query generator}. The decoder is trained to reconstruct the samples by taking the retrieved memories as input.

\par
We first define $\mbX$ to represent the domain of the data samples and $\mbZ$ to represent the domain of the encodings. 
Let $f_e(\cdot): \mbX\rightarrow\mbZ$ denote the encoder and $f_d(\cdot): \mbZ\rightarrow\mbX$ denote the decoder. 
Given a sample $\bx\in \mbX$, the encoder converts it to an encoded representation as $\bz\in \mbZ$; and the decoder is trained to reversely mapping a latent representation $\hatbz\in\mbZ$ to the domain $\mbX$ as follows
\begin{equation}
\bz=f_e(\bx; ~\theta_e),
\label{eq:encoder}
\end{equation}
\begin{equation}
\hatx=f_d(\hatbz; ~\theta_d),
\label{eq:decoder}
\end{equation}
where $\theta_e$ and $\theta_d$ denote the parameters of the encoder $f_e(\cdot)$ and decoder $f_d(\cdot)$, respectively. 
In the proposed MemAE, $\bz$ is used to retrieve the relevant memory items; and $\hatbz$ is obtained using the retrieved items. 
For the standard AE model, there is $\hatbz=\bz$. 
Our method is agnostic to the structures of the encoder and decoder, which can be specially selected for different applications.

\par
In testing, given a sample $\bx$, we use the $\ell_2$-norm based mean square error (MSE), \ie $e=\|\bx-\widehat{\bx}\|_2^2$, to measure of the reconstruction quality, which is used as the criterion for anomaly detection.

\subsection{Memory Module with Attention-based Sparse Addressing}
The proposed memory module consists of a memory to record the prototypical encoded patterns and an attention-based addressing operator for accessing the memory. 

\subsubsection{Memory-based Representation} 
The memory is designed as a matrix $\bM\in \mbR^{N\times C}$ containing $N$ real-valued vectors of fixed dimension $C$. For convenience, we assume $C$ is same to the dimension of $\bz$ and let $\mbZ=\mbR^C$.
Let the row vector $\bm_i, \forall i\in [N]$ denote the $i$-th row of $\bM$, where $[N]$ denotes the set of integers from 1 to $N$. Each $\bm_i$ denotes a memory item. 
Given a query (\ie encoding) $\bz\in \mbR^C$, the memory network obtains $\hatbz$ relying a soft \emph{addressing vector} $\bw\in\mbR^{1\times N}$ as follows
\begin{equation}
\hatbz = \bw\bM = \sum\nolimits_{i=1}^N w_i \bm_i,
\label{eq:mem_raed}
\end{equation}
where $\bw$ is a row vector with non-negative entries that sum to one and $w_i$ denotes the $i$-th entry of $\bw$. 
The weight vector $\bw$ is computed according to $\bz$. 
As shown in Eq. \eqref{eq:mem_raed}, the addressing weight $\bw$ is required for accessing the memory.
{The hyper-parameter $N$ defines the maximum capacity of the memory. 
Although it is non-trivial to find the optimal $N$ for different datasets, MemAE is insensitive to the setting of $N$, fortunately (See Section \ref{sec:exp_video}). 
A large enough $N$ can work well for each dataset.}

\subsubsection{Attention for Memory Addressing} 
\label{sec:addressing}
In MemAE, the memory $\bM$ is designed to explicitly record the prototypical normal patterns during training. We define the memory as a content addressable memory \cite{WestonCB14memnet,rae2016scaling} with an addressing scheme that computes attention weights $\bw$ based on the similarity of the memory items and the query $\bz$. As visualized in Figure \ref{fig:memae_diag}, we compute each wight $w_i$ via a softmax operation:
\vspace{-0.1cm}
\begin{equation}
w_i = \frac{\exp(d(\bz, \bm_i))}{\sum_{j=1}^{N} \exp(d(\bz, \bm_j))},
\label{eq:attention}
\vspace{-0.1cm}
\end{equation}
where $d(\cdot, \cdot)$ denotes a similarity measurement.
Similar to \cite{santoro2016one}, we define $d(\cdot, \cdot)$ as cosine similarity:
\vspace{-0.1cm}
\begin{equation}
d(\bz, \bm_i) = \frac{\bz\bm_i^\T}{\|\bz\| \|\bm_i\|}.
\label{eq:cos_dis}
\vspace{-0.1cm}
\end{equation}

\par
As shown in Eq. \eqref{eq:mem_raed}, \eqref{eq:attention} and \eqref{eq:cos_dis}, the memory module retrieves the memory items most similar to $\bz$ to obtain the representation $\hatbz$. 
Due to the restricted memory size and the sparse addressing technique (introduced in Section \ref{sec:hard_shrinkage}), only a small number of memory items can be addressed every time. Accordingly, the beneficial behaviors of the memory module can be interpreted as follows.

\par
In \emph{training phase}, the decoder in MemAE is restricted to perform reconstruction merely using a very small number of addressed memory items, rendering the requirement for efficient utilization of the memory items. 
The reconstruction supervision thus forces the memory to record the most representative prototypical patterns in the input normal patterns. 
In Figure \ref{fig:vis_mem_sampling}, we visualize the trained single memory slots, which shows that each single memory slot records the prototypical normal patterns in the training data.

\par
In \emph{testing phase}, given the trained memory, only the normal patterns in the memory can be retrieved for reconstruction. Thus the normal samples can naturally be reconstructed well. Conversely, the encoding of an abnormal input will be replaced by the retrieved normal patterns, resulting in significant reconstruction errors on anomalies (See visualized examples in Figure \ref{fig:mnist_vis}).

\subsubsection{Hard Shrinkage for Sparse Addressing}
\label{sec:hard_shrinkage}
As discussed above, performing reconstruction with a restricted number of normal patterns in the memory helps to induce large reconstruction error on anomalies. 
The attention-based addressing tends to approach this naturally \cite{graves2014neural}. 
However, some anomalies may still have the chance to be reconstructed well with a complex combination of the memory items via a dense $\bw$ containing many small elements. 
To alleviate this issue, 
we apply a hard shrinkage operation to promote the sparsity of $\bw$:
\vspace{-0.1cm}
\begin{equation}
\widehat{w}_i = h(w_i;\lambda)=\left\{
      \begin{array}{l}
        \!\!w_i,~ ~~\text{if}~~ w_i>\lambda,\\
        \!\!0, ~~~~~\text{otherwise},\\
      \end{array}
      \right.
\label{eq:shrinkage}
\vspace{-0.1cm}
\end{equation}
where $\widehat{w}_i$ denotes the $i$-th entry of the memory addressing weight vector $\widehat{\bw}$ after shrinkage and $\lambda$ denotes the shrinkage threshold. 
It is not easy to directly implement the backward of the discontinuous function in Eq. \eqref{eq:shrinkage}. 
For simplicity, considering that all entries in $\bw$ are non-negative, we rewrite the hard shrinkage operation using the continuous ReLU activation function as 
\vspace{-0.05cm}
\begin{equation}
\hatw_i = \frac{\max(w_i-\lambda, 0)\cdot w_i}{|w_i-\lambda| + \epsilon},
\label{eq:shrinkage_relu}
\vspace{-0.05cm}
\end{equation}
where $\max(\cdot, 0)$ is also known as ReLU activation, and $\epsilon$ is a very small positive scalar. 
In practice, setting the threshold $\lambda$ as a value in the interval $[1/N, 3/N]$ can render desirable results. 
After the shrinkage, we re-normalize $\hatbw$ by letting $\hatw_i=\hatw_i/\|\hatbw\|_1, \forall i$. The latent representation $\hatbz$ will be obtained via $\hatbz=\hatbw\bM$. 

\par
The sparse addressing encourages the model to represent an example using fewer but more relevant memory items, leading to learning more informative representations in memory. 
In addition, similar to the sparse representation methods \cite{zhao2011online}, encouraging sparsity of the addressing weights is beneficial in testing due to that the memory $\bM$ is trained to adapt the sparse $\bw$. Encouraging sparsity in $\bw$ will also alleviate the issue that an abnormal sample may be fairly reconstructed well with dense addressing weights. Comparing with the sparse representation methods \cite{zhao2011online,luo2017revisit}, the proposed method obtains the desired sparse $\bw$ via once efficient forward operation, instead of the iterative updating.

\subsection{Training}
Given a dataset $\{\bx^t\}_{t=1}^T$ containing $T$ samples, let $\hatx^t$ denote the reconstructed sample corresponding the each training sample $\bx^t$. 
We firstly conduct to minimize the reconstruction error on each sample:
\vspace{-0.1cm}
\begin{equation}
{R}(\bx^t,~ \hatx^t) = \|\bx^t-\hatx^t\|_2^2,
\label{eq:loss_recon_error}
\vspace{-0.1cm}
\end{equation}
where the $\ell_2$-norm is used to measure the reconstruction error. 
Let $\hatbw^t$ denote the memory addressing weights for each sample $\bx^t$. 
To further promote the sparsity of $\hatbw$, in addition to the shrinkage operation in Eq. \eqref{eq:shrinkage_relu}, we minimize a sparsity regularizer on $\hatbw$ during training. 
Considering that all entries of $\hatbw$ are non-negative and $\|\hatbw\|_1=1 $, we minimize the entropy of $\hatbw^t$:
\vspace{-0.1cm}
\begin{equation}
E(\hatbw^t) = \sum\nolimits_{i=1}^T - \hatw_i \cdot \log(\hatw_i).
\label{eq:loss_entropy}
\vspace{-0.1cm}
\end{equation}
The hard shrinkage operation in Eq. \eqref{eq:shrinkage_relu} and the entropy loss Eq. \eqref{eq:loss_entropy} jointly promote the sparsity of the generated addressing weights. 
More detailed ablation studies and discussions can be found in Section \ref{sec:exp_ablation}.
 
\par
By combining loss functions in Eq. \eqref{eq:loss_recon_error} and \eqref{eq:loss_entropy}, we construct the training objective for MemAE as:
\vspace{-0.1cm}
\begin{equation}
L(\theta_e, \theta_d, \bM) = \frac{1}{T}\sum\nolimits_{t=1}^T \left(R(\bx^t,~ \hatx^t) + \alpha E(\hatbw^t) \right),
\label{eq:loss_full}
\vspace{-0.1cm}
\end{equation}
where $\alpha$ is a hyper-parameter in training. In practice, $\alpha=0.0002$ leads to desirable results in all our experiments. 
During training, the memory $\bM$ is updated through optimization via backpropagation and gradient descent. In backward pass, only the gradients for the memory items with non-zero addressing weights $w_i$ can be non-zero.

\section{Experiments}
In this section, we validate the proposed MemAE for anomaly detection. 
To show the generality and applicability of the proposed model, we conduct experiments on five datasets of three different tasks. 
The results are compared with different baseline models and state-of-the-art techniques. 
The proposed MemAE is applied to all datasets following previous sections. 
MemAE and its variants are implemented using PyTorch \cite{paszke2017automatic} and trained using the optimizer Adam \cite{kingma2014adam} with a learning rate of 0.0001. We make them and other encoder-decoder models such as VAE to have the similar model capacity.

\subsection{Experiments on Image Data}
\label{sec:exp_img}
We first conduct the experiments to detect outliers image \cite{sabokrou2018adversarially} and evaluate the performance on two image datasets: MNIST \cite{lecun1998mnist} and CIFAR-10 \cite{krizhevsky2009learning}, both of which contain images belonging to 10 classes. For each dataset, we construct 10 anomaly detection (\ie one-class classification) datasets by sampling images from each class as normal samples and sampling anomalies from the rest classes. The normal datums are split into training and testing set with a rate of 2:1. 
Following the setting used in \cite{zhai2016deep,zong2018deep}, the training set only consists of normal samples and has no overlapping with the testing set. The anomaly proposition is controlled around 30\%. 10\% of the original training data is left for validation.

\par
In this experiment, we focus on validating the proposed memory module and implement the encoder and decoder as plain convolutional neural networks. 
We first define $\text{Conv2}(k, s, c)$ to denote a 2D convolution layer, where $k$, $s$ and $c$ are the kernel size, stride size and the number of kernels, respectively. For MNIST, we implement the encoder using three convolution layers: {Conv2}(1, 2, 16)-{Conv2}(3, 2, 32)-{Conv2}(3, 2, 64).  
The decoder is implemented as {Dconv2}(3, 2, 64)-{Dconv2}(3, 2, 32)-{Dconv2}(3, 2, 1), where Dconv2 denotes the 2D deconvolution layer. Except for the last Dconv2, each layer is followed by a batch normalization (BN) \cite{ioffe2015batch} and a ReLU activation \cite{nair2010rectified}. Such design is applied for all datasets in the following. 
Considering the higher data complexity of CIFAR-10, we use the encoder and decoder with larger capacities: {Conv2}(3, 2, 64)-{Conv2}(3, 2, 128)-Conv2(3, 2, 128)-Conv2(3, 2, 256) and Dconv2(3, 2, 256)-Dconv2(3, 2, 128)-Dconv2(3, 2, 128)-Dconv2(3, 2, 3). 
We process the MNIST and CIFAR-10 datasets as gray images and RGB images, respectively. 
Memory sizes $N$ for MNIST and CIFAR-10 are set as 100 and 500, respectively.

\par
We compare the proposed model with several conventional and deep learning based methods for general anomaly detection as baselines, including one-class SVM (OC-SVM) \cite{scholkopf2000support}, kernel density estimation (KDE) \cite{parzen1962estimation}, a deep variational autoencoder (VAE) \cite{kingma2013auto}, a deep autoregressive generative model PixCNN \cite{van2016conditional} and the deep structured energy-based model (DSEBM) \cite{zhai2016deep}. Specifically, for the density estimation methods (\eg KDE and PixCNN) and the reconstruction based methods (\eg VAE and DSEBM), the log-likelihood and reconstruction error are used to calculate the regularity score, respectively. Note that for fair comparison with other methods, we calculate the regularity score of VAE based on only reconstruction error.
We also conduct the comparisons with some baseline variants of MemAE to show the importance of the major components, including the antueocoder without memory module (AE) and a variant of MemAE without the sparse shrinkage and the entropy loss (MemAE-nonSpar). In all experiments, AE, MemAE-nonSpar, and VAE share a similar capacity with the full MemAE model by using the same encoder and decoder. In testing, we scale the reconstruction error to the range $[0,1]$ as the criterion to identify the anomalies. Following \cite{mahadevan2010anomaly,luo2017revisit,abati2018and}, we use the AUC (Area Under Curve) as the measurement for performance evaluation, which is obtained by calculating the area under the Receiver Operation Characteristic (ROC) with a varying threshold. 
Table \ref{tab:image_data} shows the {average} AUC values on the 10 sampled datasets.

\begin{table}[!t]
\begin{center}
\caption{Experimental results on image data. Average AUC values on 10 anomaly detection datasets sampled from MNIST and CIFAR-10 are shown. }
\label{tab:image_data}
\small
\begin{tabular}{l|cc}
\hline
Dataset          & ~~~~MNIST~ & ~CIFAR-10~ \\ \hline
OC-SVM \cite{scholkopf2000support}       & ~~~~~0.9499~      &  ~~0.5619~        \\ KDE           & ~~~~~0.8116~      &  ~~0.5756~        \\ VAE \cite{kingma2013auto}          & ~~~~~0.9643~      &  ~~0.5725~       \\ PixCNN \cite{van2016conditional}       & ~~~~~0.6141~      &  ~~0.5450~        \\ DSEBM \cite{zhai2016deep}        & ~~~~~0.9554~      &  ~~0.5725~        \\ \hline
AE            & ~~~~~0.9619~      &  ~~0.5706~        \\ MemAE-nonSpar~ & ~~~~~0.9725~      &  ~~0.6058~        \\ MemAE         & ~~~~~\bf 0.9751~      &  ~~\bf 0.6088~       \\ \hline
\end{tabular}
\vspace{-0.7cm}
\end{center}
\end{table}

\par
As shown in Table \ref{tab:image_data}, the proposed MemAE generally outperforms the compared methods. The memory-augmented models significantly outperform the AE without memory. And the MemAE model with sparse addressing yields better results. 
Images in MNIST only contain simple patterns, \ie digits, which is easy to model. VAE can thus produce satisfactory results by using a simple Gaussian distribution to model latent space. 
All methods perform better on MNIST than CIFAR-10, since the images in CIFAR-10 have more complex content and exhibit larger intra-class variance on several classes, which incurs unsatisfactory average ACU. 
Nevertheless, among the compared models with similar capacities, MemAE achieves superior performance than the competitors, which proves the effectiveness of the proposed memory module.

\subsubsection{Visualizing How the Memory Works}
\label{sec:visl_memory}
Considering that the images in MNIST contain the patterns easy to identify, we use it to show how the proposed memory module works for anomaly detection.

\par
\begin{figure}[htp]
\vspace{-0.1cm}
\centering
\subfigure[Training samples]{
\centering
\includegraphics[width=0.07\textwidth]{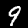}
\includegraphics[width=0.07\textwidth]{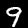}
\label{fig:tr_sample_9}
}
\subfigure[Decoded single memory item]{
\centering
\includegraphics[width=0.07\textwidth]{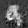}
\includegraphics[width=0.07\textwidth]{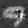}
\includegraphics[width=0.07\textwidth]{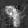}
\includegraphics[width=0.07\textwidth]{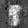}
\label{fig:mem_slots_vis}
}
\caption{Visualization of the memory slots learned on MNIST by treating digits ``9'' as normal data. We randomly select a single memory item and perform decoding.
 The decoded single memory slot in (b) appears as a prototypical pattern of the normal samples.}
\label{fig:vis_mem_sampling}
\end{figure}

\noindent \textbf{What the memory learns.} 
We first visualize what the memory learns from MNIST by randomly sampling a single memory slot and performing decoding on it. Figure \ref{fig:vis_mem_sampling} visualizes the memory learned on the MNIST digits ``9'' by treating them as normal samples. 
Since MemAE usually performs reconstruction via a combination of several addressed items, the decoded single slot appears blurry and noisy. Nevertheless, as shown in Figure \ref{fig:mem_slots_vis}, the memory slots record the different prototypical patterns of the normal training samples (\ie digits ``9'').

\par
\begin{figure}[!t]
\vspace{-0.1cm}
\centering
\subfigure[Training on the normal ``5'']{
\centering
\begin{overpic}[trim =10mm 10mm 0mm 0mm, clip,
width=0.15\linewidth]
{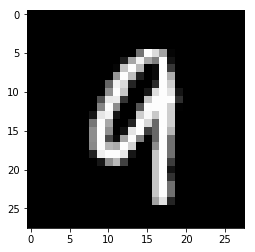}
\put(2,3){\scriptsize\color{white}{\bf Input}}
\end{overpic}
\begin{overpic}[trim =10mm 10mm 0mm 0mm, clip,
width=0.15\linewidth]
{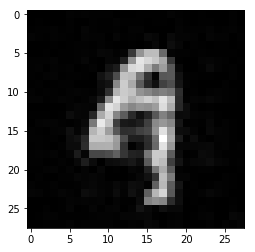}
\put(2,3){\scriptsize\color{white}{\bf AE}}
\end{overpic}
\begin{overpic}[trim =10mm 10mm 0mm 0mm, clip,
width=0.15\linewidth]
{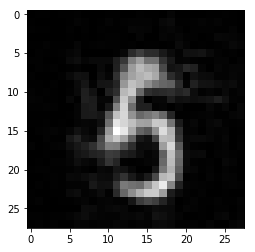}
\put(2,3){\scriptsize\color{white}{\bf MemAE}}
\end{overpic}
\label{fig:tr_on5}
}
\subfigure[Training on the normal ``2'']{
\label{fig:psnr_vs_ite_bsd}
\centering
\begin{overpic}[trim =10mm 10mm 0mm 0mm, clip,
width=0.15\linewidth]
{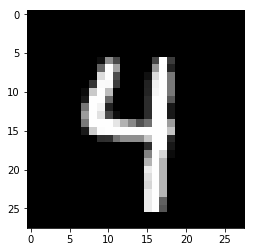}
\put(2,3){\scriptsize\color{white}{\bf Input}}
\end{overpic}
\begin{overpic}[trim =10mm 10mm 0mm 0mm, clip,
width=0.15\linewidth]
{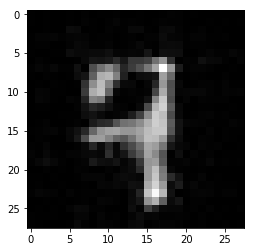}
\put(2,3){\scriptsize\color{white}{\bf AE}}
\end{overpic}
\begin{overpic}[trim =10mm 10mm 0mm 0mm, clip,
width=0.15\linewidth]
{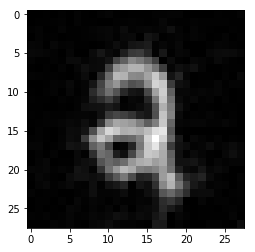}
\put(2,3){\scriptsize\color{white}{\bf MemAE}}
\end{overpic}
}
\caption{Visualization of the reconstruction results of AE and MemAE on MNIST. (a) The models are trained on ``5''. The input is an image of ``9''. (b) The models are trained on ``2''. The input is an image of ``4''. The MemAE retrieves the normal memory items for reconstructions and obtains the results significantly different from the input anomalies.}
\label{fig:mnist_vis}\vspace{-0.1cm}
\end{figure}

\noindent \textbf{How memory augments reconstruction.}
In Figure \ref{fig:mnist_vis}, we visualize the image reconstruction process under the memory augmentation. 
Since the trained memory only records the normal prototypical patterns, given an abnormal input ``9'', the MemAE trained on ``5'' reconstructs a ``5'', resulting in significant reconstruction error on the abnormal input. Note that the reconstructed ``5'' of MemAE has a similar shape of the input ``9'' since the memory module retrieves the most similar normal patterns. The AE model without memory tends to learn some representations more locally. Thus an abnormal sample may also be reconstructed well.

\subsection{Experiments on Video Anomaly Detection}
\label{sec:exp_video}
Anomaly detection on video aims to identify the unusual contents and moving patterns in the video, which is an essential task in video surveillance. 
We conduct experiments on three real-world video anomaly detection datasets, \ie UCSD-Ped2 \cite{mahadevan2010anomaly}, CUHK Avenue \cite{lu2013abnormal} and ShanghaiTech \cite{luo2017revisit}. Specifically, the most recent benchmark dataset ShanghaiTech contains more than 270,000 training frames and more than 42,000 frames (with about 17,000 abnormal frames) for testing, which covers 13 different scenes. 
In the datasets, objects except for pedestrians (\eg vehicles) and strenuous motion (\eg fighting and chasing) are treated as anomalies.

\begin{figure}[!t]
\vspace{-0.1cm}
\centering
\subfigure[UCSD-Ped2]{
\centering
\includegraphics[trim=27 0 35 17, clip, width=0.22\textwidth]{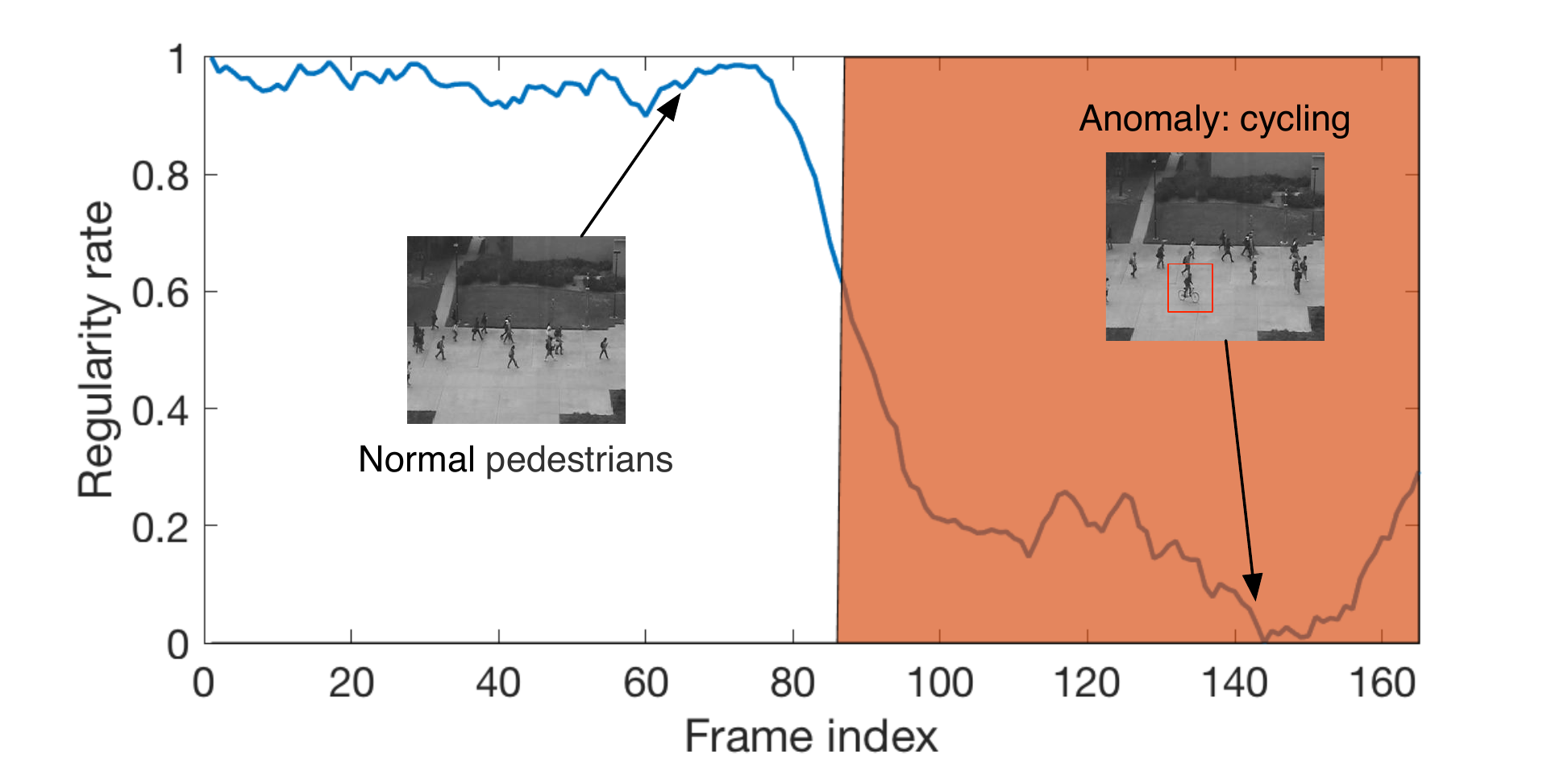}
\label{fig:ucsd_score}
}
\subfigure[ShanghaiTech]{
\centering
\includegraphics[trim=27 0 35 17, clip, width=0.22\textwidth]{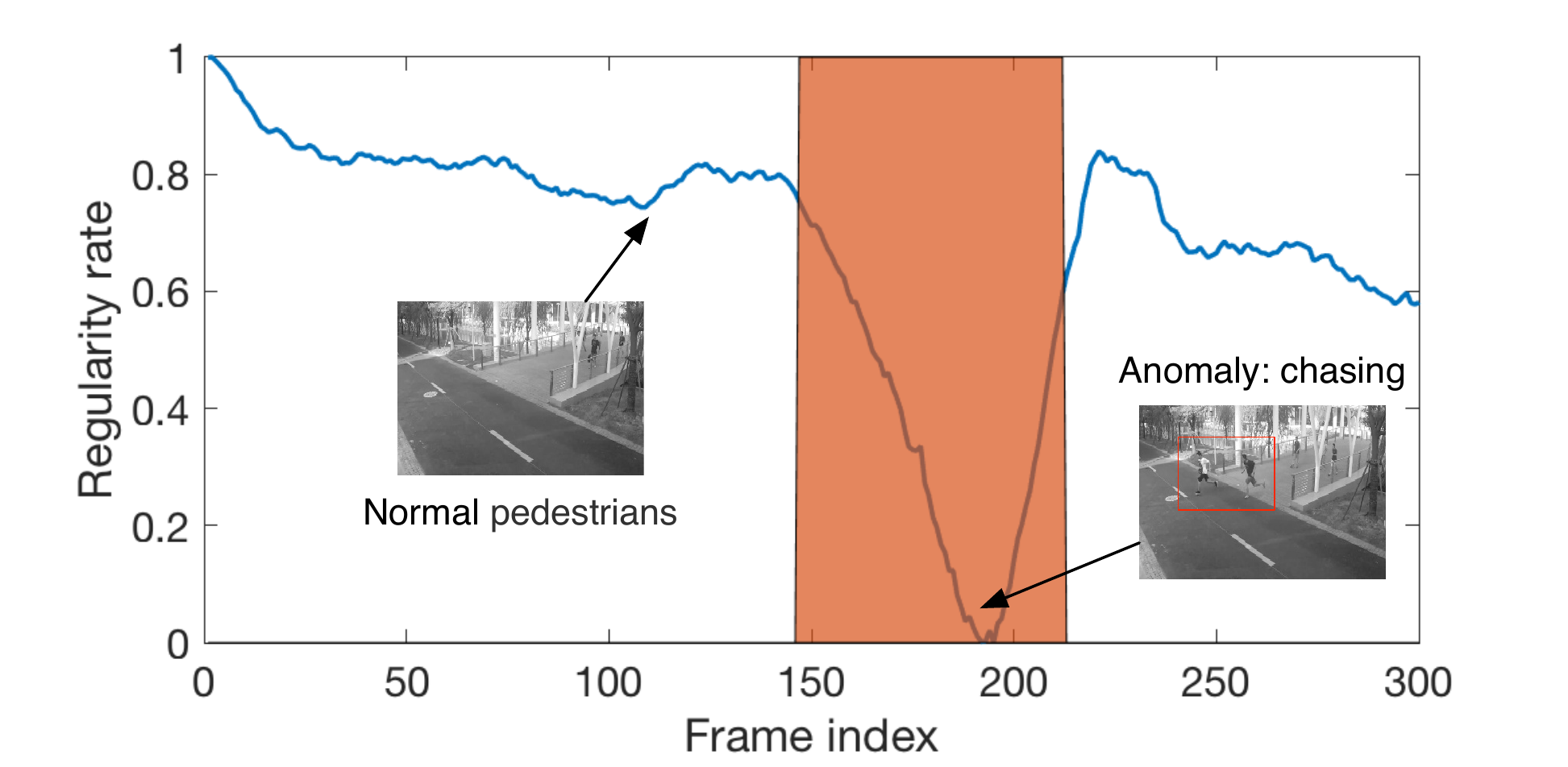}
\label{fig:sht_score}
}
\caption{Normality scores of the video frames obtain by MemAE. The score decreases immediately when some anomalies appear in the video frame.}
\vspace{-0.3cm}
\label{fig:video_score}
\end{figure}

\par
To preserve the video temporal information, we implement the encoder and decoder using 3D convolutions to extract the spatial-temporal features in video \cite{tran2015learning}. Accordingly, the input of the network is a cuboid constructed by stacking 16 neighbor frames in grayscale. The structures of encoder and decoder are designed as: {Conv3}(3, 2, 96)-Conv3(3, 2, 128)-Conv3(3, 2, 256)-Conv3(3, 2, 256) and Dconv3(3, 2, 256)-Dconv3(3, 2, 256)-Dconv3(3, 2, 128)-Dconv3(3, 2, 1), where Conv3 and Dconv3 denote 3D convolution and deconvolution, respectively. A BN and a ReLU activation follow each layer (except the last one).
We set $N=2000$. 
Considering the complexity of the video data, we let each memory slot record the features on one pixel in the feature maps, corresponding to a sub-area of the video clip. The memory is thus a matrix of $2000\times 256$. 
In testing, the normality of each frame is evaluated by the reconstruction error of the cuboid centering on it. Following \cite{hasan2016learning,luo2017revisit}, we obtain the normality score $p_u$ of the $u$-th frame by normalizing the errors to range $[0,1]$:
\vspace{-0.1cm}
\begin{equation}
p_u = 1-\frac{e_u-\min_u(e_u)}{\max_u(e_u)-\min_u(e_u)},
\vspace{-0.1cm}
\end{equation}
where $e_u$ denotes the reconstruction error the $u$-th frame in a video episode. The value of $p_u$ closer to $0$ indicates the frame is more likely an abnormal frame. 
Figure \ref{fig:video_score} shows that the normality score obtained by MemAE immediately decreases when some anomalies appear in the video frame.

\begin{table}[!t]
\begin{center}
\caption{AUC of different methods on video datasets UCSD-Ped2, CUHK Avenue and ShanghaiTech.}
\label{tab:video_data}
\small
\begin{tabular}{c|c|ccc}
\hline
\multicolumn{2}{c|}{Method\textbackslash{}Dataset} & {UCSD-Ped2} & CUHK & {SH.Tech} \\ \hline
\multirow{6}{*}{\rotatebox[origin=c]{90}{Non-Recon.}} & MPPCA \cite{kim2009observe}            & 0.693~& - & ~-~ \\
& MPPCA+SFA \cite{mahadevan2010anomaly}  & 0.613~& - & ~-~  \\
& MDT \cite{mahadevan2010anomaly}        & 0.829~& - & ~-~  \\
& AMDN \cite{xu2015learning}             & 0.908~& - & ~-~  \\
& Unmasking \cite{tudor2017unmasking}    & 0.822~& 0.806 & ~-~ \\ 
& MT-FRCN \cite{hinami2017joint}         & 0.922~& - & ~-~ \\ 
& Frame-Pred \cite{luo2017revisit}       & \bf 0.954~& \bf 0.849 & \bf~0.728~ \\ \hline \multirow{8}{*}{\rotatebox[origin=c]{90}{Recon.}} & AE-Conv2D \cite{hasan2016learning}     & 0.850~& 0.800 & ~0.609~  \\
& AE-Conv3D \cite{zhao2017spatio}        & 0.912~& 0.771 & ~-~ \\
& TSC         \cite{luo2017revisit}      & 0.910~& 0.806 & ~0.679~ \\
& StackRNN \cite{luo2017revisit}      & 0.922~& 0.817 & ~0.680~ \\ \cline{2-5} 
& AE                                     & 0.917~& 0.810 & ~0.697~ \\
& MemAE-nonSpar                          & 0.929~& 0.821 & ~0.688~ \\
& MemAE                                  & \bf 0.941~& \bf 0.833 &  ~\bf 0.712~  \\ \hline
\end{tabular}
\vspace{-0.6cm}
\end{center}
\end{table}

\par
Due to the complexity of the video data, many general anomaly detection methods \cite{parzen1962estimation,kingma2013auto,zong2018deep} without specific design cannot work well on videos. 
To show the effectiveness of the proposed memory module, we compare the proposed MemAE with many well-designed reconstruction based state-of-the-art methods including AE methods with 2D \cite{hasan2016learning} and 3D convolution \cite{zhao2017spatio} (AE-Conv2D and AE-Conv3D), 
a temporally-coherent sparse coding method (TST) \cite{luo2017revisit}, a stacked recurrent neural network (StackRNN) \cite{luo2017revisit} and many video anomaly detection baselines. 
The variants of MemAE are also compared as baselines.

\par
\begin{figure}[!t]
\centering
\subfigure[Frame]{
\centering
\includegraphics[trim=10 5 0 60, clip, width=0.12\textwidth]{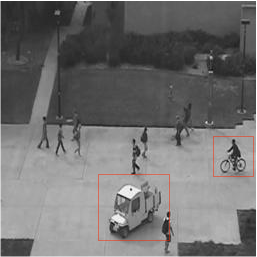}
\label{fig:vis_recon_error_frame}
}
\subfigure[AE]{
\centering
\includegraphics[trim=10 5 0 60, clip, width=0.12\textwidth]{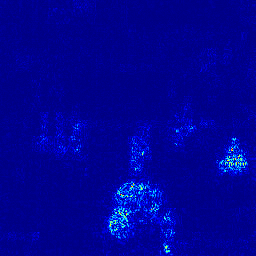}
\label{fig:vis_recon_error_ae}
}
\subfigure[MemAE]{
\centering
\includegraphics[trim=10 5 0 60, clip, width=0.12\textwidth]{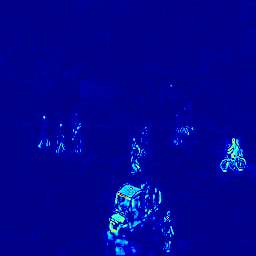}
\label{fig:vis_recon_error_ae}
}
\caption{Reconstruction error of AE and MemAE on an abnormal frame of UCSD-Ped2. {MemAE can significantly highlight the abnormal parts (in \redtext{red} bounding box) in the scene.}}
\label{fig:recon_error_map}
\vspace{-0.3cm}
\end{figure}

Table \ref{tab:video_data} shows the AUC values on video datasets. 
MemAE produces much better results than TSC and StackRNN \cite{luo2017revisit}, which also apply the sparse regularization.
The comparisons with AE and MemAE-nonSpar show that the memory module with the sparse addressing is steadily beneficial.
Figure \ref{fig:recon_error_map} visualizes the reconstruction error on one abnormal frame in UCSD-Ped2. The error map of MemAE significantly highlights the abnormal event (\ie vehicle and bicycle moving on the sidewalk), inducing low normality score. However, AE reconstructs the anomaly well and induces some random errors. 

The proposed MemAE obtains better or comparative performance than other methods, while our model solves a more general problem and can be flexibly applied to different types of data. 
By merely using the reconstruction error, the proposed method can obtain superior results with the minimum knowledge of the specific application. 
Even comparing with the method \cite{liu2017future}  (\ie Frame-Pred in Table \ref{tab:video_data}) that uses many non-reconstruction techniques specifically for video data, \eg optical flow, frame prediction, and adversarial loss, the performance of the proposed MemAE is still comparable.  
Note that the purpose of our experiment is not pursuing the highest accuracy on certain applications but to demonstrate the advantages of the proposed improvement of AE, \ie MemAE, for the general anomaly detection problem.
Our study is orthogonal to that in \cite{liu2017future} and can be readily incorporated into their system to boost the performance further. On the other hand, the techniques in \cite{liu2017future} can also be used in the proposed MemAE.

\par
\begin{figure}[htp]
\vspace{-0.1cm}
\centering
\includegraphics[trim=0 0 0 5, clip, width=0.4\textwidth]{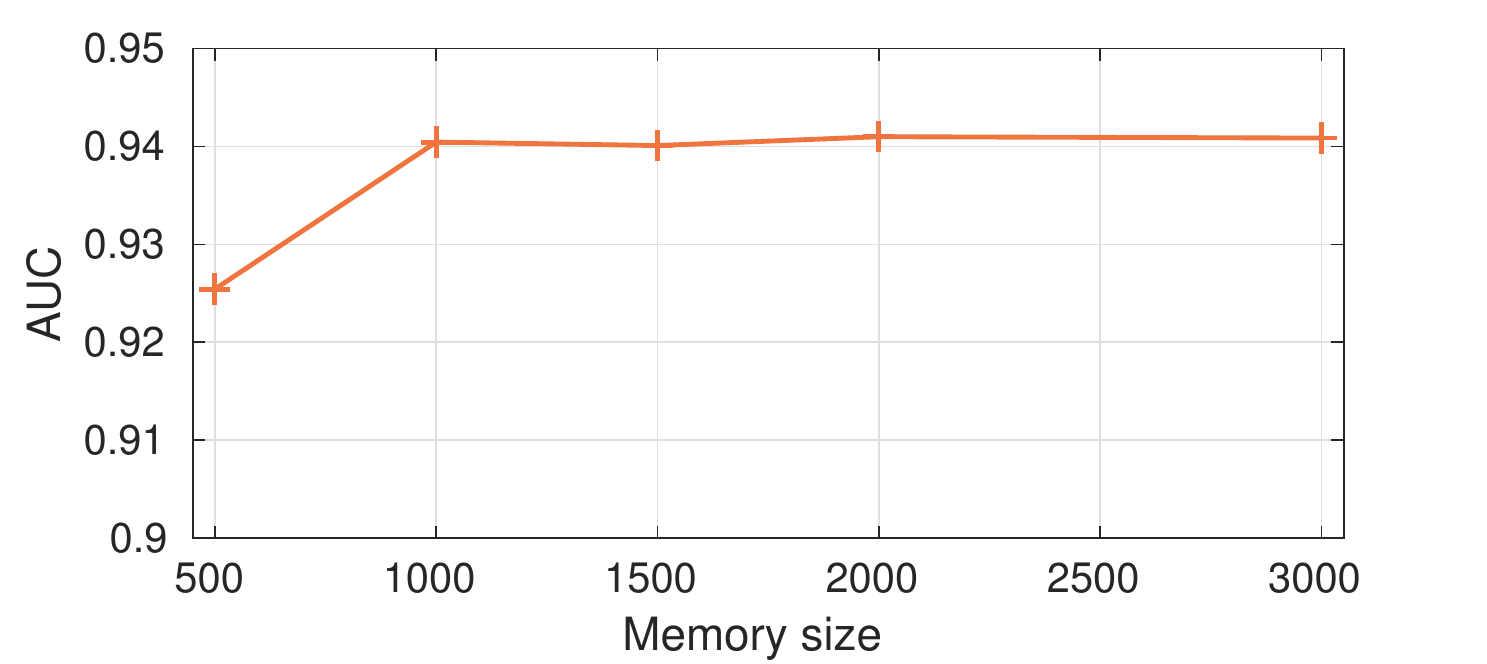}
\caption{Robustness to the setting of memory size. AUC values of MemAE with different memory size on UCSD-Ped2 are shown. }
\label{fig:memory_size_robust}
\end{figure}

\noindent \textbf{Robustness to the memory size}~ We use the UCSD-Ped2 to study the robustness of the proposed MemAE to the memory size $N$. We conduct the experiments by using different memory size settings and show the AUC values in Figure \ref{fig:memory_size_robust}. Given a large enough memory size, the MemAE can robustly produce plausible results.

\vspace{-0.1cm}

\par
\noindent \textbf{Running time}~ 
We empirically study the computing complexity of the proposed method on the video dataset UCSD-Ped2 with an NVIDIA GeForce 1080 Ti graphics card. The proposed MemAE averagely takes 0.0262 seconds for video anomaly detection of one frame (\ie 38 fps), which is on par or faster than previous state-of-the-art deep learning based methods such as \cite{liu2017future} using 0.04s, \cite{luo2017revisit} using 0.02s and \cite{tudor2017unmasking} using 0.05s\footnote{Running time of the compared methods are quoted from \cite{liu2017future} for reference, which is produced using a faster graphics card than ours.}. Moreover, comparing to our baseline AE model that takes 0.0266s for each frame, our memory module (in MemAE) induces little additional computation time (\ie $4\!\times \!\!10^{-4}$s per frame).

\subsection{Experiments on Cybersecurity Data}
To further validate the generalization of the proposed method, we experiment on a widely used cybersecurity dataset {beyond the computer vision applications}, \ie KDDCUP99 10 percent dataset from the UCI repository \cite{lichman2013uci}.
Following the settings in \cite{zong2018deep}, $80\%$ of the samples labeled as ``attack'' in the original dataset are treated as normal samples. Each sample can be organized as a vector with 120 dimensions \cite{zong2018deep}. 
We use fully-connected layers (noted as FC) to implement the encoder and decoder as FC(120, 60)-FC(60, 30)-FC(30, 10)-FC(10, 3) and FC(3, 10)-FC(10, 30)-FC(30, 60)-FC(60, 120), in which FC($i$, $o$) denotes the FC layer with input and output size $i$ and $o$. 
Expect the last one, each FC layer is followed by a Tanh activation. The structure shares a similar capacity to the model in \cite{zong2018deep}. We set $N=50$ and thus have a memory with the size of $50\times 3$.

\par
As suggested in \cite{zhai2016deep,zong2018deep}, we randomly sample $50\%$ of data for training and the rest for testing. Only data samples from normal class are used for training. We compare the proposed method with previous state-of-the-art methods on the KDDCUP dataset, including OC-SVM \cite{scholkopf2000support}, a deep clustering network (DCN) \cite{yang2017joint}, DSEBM \cite{zhai2016deep}, DAGMM \cite{zong2018deep} and the baseline variants of MemAE. 
Following the standard protocol \cite{zong2018deep}, the methods are evaluated using the average precision, recall and $F_1$ score after 20 runs. 
DAGMM and the proposed models perform very well because of the more effective data modeling. The proposed method obtains the superior performance 
since it can explicitly memorize the behavior patterns of ``attack'' samples.

\begin{table}[!t]
\begin{center}
\caption{Experimental results of different methods on the cybersecurity dataset KDDCUP.}
\label{tab:cyber_data}
\small
\begin{tabular}{l|ccc}
\hline
Method\textbackslash{}Metric          & Precision & Recall & $F_1$  \\ \hline
OC-SVM \cite{scholkopf2000support} & ~~0.7457&~~0.8523 &~~0.7954 \\
DCN \cite{yang2017joint} & ~~0.7696 & ~~0.7829 &~~0.7762 \\
DSEBM \cite{zhai2016deep} & ~~0.8619 & ~~0.6446 & ~~0.7399 \\
DAGMM \cite{zong2018deep}         & ~~0.9297  & ~~0.9442 & ~~0.9369\\ \hline
AE            &   ~~0.9328     &  ~~0.9356     &  ~~0.9342 \\ MemAE-nonSpar~ & ~~0.9341 & ~~0.9368   & ~~0.9355 \\ MemAE         & ~~\bf 0.9627  & ~~\bf  0.9655    & ~~\bf 0.9641  \\ \hline
\end{tabular}
\vspace{-0.8cm}
\end{center}
\end{table}

\subsection{Ablation Studies}
\label{sec:exp_ablation}
In previous sections, extensive comparisons among MemAE and its variants, \ie AE and MemAE-nonSpar, have proved the importance of the major components of the proposed method. In this section, we will conduct several further ablation studies to investigate other different components in details.

\subsubsection{Study of the Sparsity-inducing Components} 
As introduced above, we use two components to induce sparsity of the memory addressing weights, \ie the hard-thresholding shrinkage defined in Eq. \eqref{eq:shrinkage} and the entropy loss $E(\cdot)$ in Eq. \eqref{eq:loss_full}. We experiment to study the importance of each component by removing the other one. Table \ref{tab:abl_sparse} records the AUC on the dataset UCSD-Ped2. As shown in Table \ref{tab:abl_sparse}, removing either the shrinkage operator or the entropy loss will degenerate the performance. 
Without the hard shrinkage, the model cannot directly encourage sparsity in testing, which may lead to non-sparse memory addressing weights with too much noise. 
Entropy loss plays a vital role when the under-trained model generates unoptimized addressing weights at the early stage of training.

\begin{table}[!t]
\begin{center}
\caption{Ablation studies based on UCSD-Ped2 dataset.}
\label{tab:abl_sparse}
\small
\begin{tabular}{l|c}
\hline
Method            & ~~~AUC~~~~\\ \hline
AE            & ~~0.9170~~\\ AE-$\ell_1$~ & ~~0.9286~~\\ \hline
MemAE-nonSpar~ & ~~0.9293~~\\ MemAE w/o Shrinkage & ~~0.9324~~\\
MemAE w/o Entropy loss~ & ~~0.9372~~\\ \hline
MemAE         & ~~\bf 0.9410~~\\ \hline
\end{tabular}
\end{center}
\vspace{-0.7cm}
\end{table}

\subsubsection{Comparison with AE with Sparse Regularization} 
The sparse memory addressing in MemAE derives a flavor of the autoencoders that induce sparsity of the encoder output (activations). We thus conduct a straightforward experiment to compare MemAE with an autoencoder with sparse regularization on the encoded features, which is directly implemented by minimizing the $\ell_1$-norm of the latent compressed feature, \ie $\|\bz\|_1$, during training, referred to as AE-$\ell_1$, which shares the same encoder and decoder with MemAE. As shown in Table \ref{tab:abl_sparse}, the performance of AE-$\ell_1$ is close to MemAE-nonSpar and superior to AE due to the sparsity-inducing regularization. However, AE-$\ell_1$ still lacks a clear mechanism to encourage large reconstruction errors on anomalies or a powerful module to model the prototypical patterns of the normal samples, lead to worse performance than MemAE and other MemAE variants.

\section{Conclusion}
In this paper, we proposed a memory-augmented autoencoder (MemAE) to improve the performance of the autoencoder based unsupervised anomaly detection methods. 
Given an input, the propose MemAE first uses the encoder to obtain an encoded representation and then use the encoding as a query to retrieve the most relevant patterns in the memory for reconstruction. Since the memory is trained to record the prototypical normal patterns, the proposed MemAE can well reconstruct the normal samples and enlarge the reconstruction error of the anomalies, which strengths the reconstruction error as the anomaly detection criterion. Experiments on various datasets from different applications prove the generalization and effectiveness of the proposed method. 
In the future, we will investigate to use the addressing weight for anomaly detection. 
Considering that the proposed memory module is general and agnostic to the structures of the encoder and decoder, we will integrate it into more complicated base models and apply it on more challenging applications.

{\small
\bibliographystyle{ieee}
\bibliography{anomdec}
}
\end{document}

%% file: dg_usepackage.tex
\usepackage{times}
\usepackage{epsfig}
\usepackage{graphicx}

\usepackage{amsmath}
\usepackage{amssymb}
\usepackage{bm}
\usepackage{subfigure}
\usepackage{multirow}%
\usepackage{color}
\usepackage{upgreek,xspace}

\usepackage{helvet}
\usepackage{courier}
\usepackage{bm}
\usepackage{algorithmic}
\usepackage{graphicx}
\usepackage{subfigure}
\usepackage{multirow}%
\usepackage{amsmath, amssymb}
\usepackage{color}
\usepackage{mathrsfs}
\usepackage{amsthm}
\usepackage{epstopdf}
\usepackage{wrapfig}
\usepackage{picinpar}
\usepackage{threeparttable, eucal}
\usepackage{enumitem}
\usepackage{subfigure}
\usepackage[vlined,ruled,linesnumbered]{algorithm2e}
\usepackage{overpic}

\usepackage{epstopdf}
\usepackage{array}

\usepackage{url}
\usepackage[usenames,dvipsnames]{xcolor}

%% file: dg_def.tex
\def\bw{{\bf w}}
\def\bx{{\bf x}}

\def\bz{{\bf z}}
\def\bm{{\bf m}}

\def\hatw{{\widehat w}}
\def\hatbw{{\widehat{\bf w}}}
\def\hatbz{{\widehat{\bf z}}}

\def\0{{\bf 0}}
\def\1{{\bf 1}}

\def\bM{{\bf M}}

\def\mbR{{\mathbb R}}
\def\mbX{{\mathbb X}}
\def\mbZ{{\mathbb Z}}

\def\hatx{\widehat{\bx}}

\def\etal{\emph{et al. }}
\def\ie{\emph{i.e. }}
\def\eg{\emph{e.g. }}

\def\T{{\sf T}}